# Region based Ensemble Learning Network for Fine-grained Classification

Weikuang Li, Tian Wang*, Chuanyun Wang, Guangcun Shan*, Mengyi Zhang, Hichem Snoussi
*School of Computer Science Shenyang Aerospace University* Shenyang, China wangcy0301@sau.edu.cn
*School of Automation Science and Electrical Engineering, Beihang University, Beijing, China*
*School of Instrumentation Science and Opto-electronics Engineering, Beihang University, Beijing, China*
*College of Electrical Engineering and Control Science, Nanjing Tech University* Nanjing, China;
*Institue Charles Delaunay-LM2S-UMR STMR 6279 CNRS, University of Technology of Troyes* Troyes, France

\* *Email: wangtian@buaa.edu.cn, or* gcshan@buaa.edu.cn

*Abstract:* As an important research topic in computer vision, fine-grained classification which aims to recognition subordinate-level categories has attracted significant attention. We propose a novel region based ensemble learning network for fine-grained classification. Our approach contains a detection module and a module for classification. The detection module is based on the faster R-CNN framework to locate the semantic regions of the object. The classification module using an ensemble learning method, which trains a set of sub-classifiers for different semantic regions and combines them together to get a stronger classifier. In the evaluation, we implement experiments on the CUB-2011 dataset and the result of experiments proves our method's efficient for fine-grained classification. We also extend our approach to remote scene recognition and evaluate it on the NWPU-RESISC45 dataset.

Keywords: **Fine-grained classification, Ensemble learning, Region detection, Remote sensing image**

## I. Introduction

The visual fine-grained classification problem is an im- portant and basic problem in computer vision. Fine-grained classification aims to distinguishing subordinate-level cate- gories belonging to the same basic-level category, such as different kinds of aircraft models [1], cars [2] [3],birds [4],dogs [5],flowers [6] and so on. Compared with generic object clas- sification, whose task is to recognition basic-level categories (e.g. bird, horse, airplane, automobile, and so on in cifar- 10 [7]), fine-grained classification is extremely challenging. Because in fine-grained classification the most parts of dif- ferent categories are similar or same and subtle difference only appears in some small semantic regions. Zhang *et al.* [8] propose a part based R-CNNs for fine-grained classification using deep convolutional features computed on bottom-up region proposals. Huang *et al.* [9] use a fully convolutional network for object regions localization, use a two-stream network to simultaneously encode object-level and part-level regions and then stack them together for recognition.

We propose a region based ensemble learning network which combines a set of region based classifiers and getting a stronger classifier. With different semantic regions located by the faster R-CNN [10] based detector, different sub-classifiers are trained. All the sub-classifiers' predictions are combined by a plurality voting strategy and a stronger classifier is got. As an efficient method in machine learning, the sub-classifiers' diversity is the essence for ensemble learning. Several methods use re-weighting strategy which changes dataset distribution to train different classifiers, such as adaboost [11]. Other methods re-sample the origin dataset to train different sub-classifiers, such as bagging [12] ,random forest [13]. Those methods train different sub-classifiers by changing the origin dataset. Our method uses different semantic regions(such as bird's tail, bird's wing, bird's back in Caltech-UCSD birds dataset [4]) to train different sub-classifiers. This makes sub-classifiers contain diversity naturally.

In experiments, we test our method on CUB-2011 dataset [4], a widely used dataset for fine-grained classification, and get a good result. We also test our method on the NWPU- RESISC45 dataset [14] and prove it can be used for remote scene classification.

## II. Approcah

This section presents the framework of the proposed region based ensemble learning network. As illustrated in Fig. 1, The framework contains two modules. The first module is a detector for the semantic region localization and the second module is a classifier for fine-grained classification.

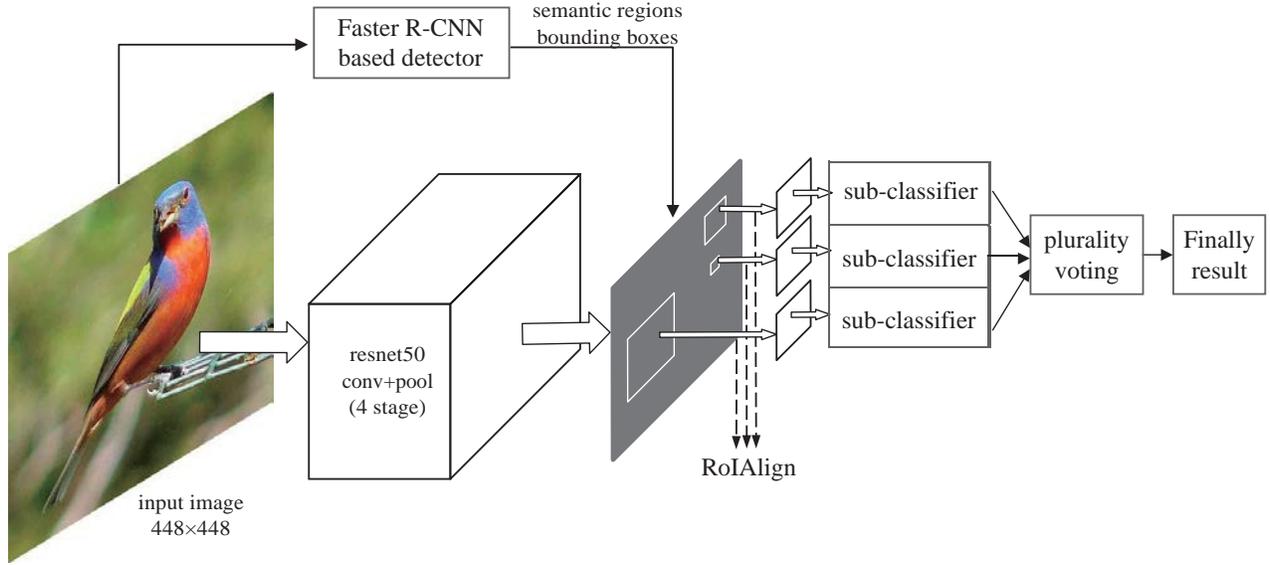

Fig. 1. The architecture of the region based ensemble learning network.

*A. Faster R-CNN based detector for localization*

The first module of the proposed framework is a detector which is used to localize the object's semantic regions. We implement the state-of-art object detection algorithm based on faster R-CNN [10] for the semantic part localization. As the two-stage detection framework, region proposals are firstly generated by the RPN based on anchors and then each proposal is classified and regressed by the second stage network to get the final detection results. In our particular, ROIAlign in the mask R-CNN [15] is used instead of ROIPooling in the origin faster R-CNN.

As a post-processing approach in faster R-CNN, Non- Maximal Suppression (NMS) will reject a region if it has an Intersection-over-Union (IoU) overlap with a higher scoring selected region larger than a pre-set threshold. NMS can fix the multiple detections problem and thus will improve the average precision. However, prior knowledge that in the semantic region localization each semantic region appears at most once is not considered in NMS. Considering this prior, we use a special NMS which will only keep the region with the highest scoring and discard others. This method should improve the detection average precise because it can reduce the false positive by fixing the multiple detections problem more reasonable.

*B. Ensemble Learning Classifier*

The second module of the framework is an ensemble learning classifier. We propose a region-based ensemble learning methods. For each kind of semantic regions, a set of sub-classifiers are trained and then are combined to get a stronger classifier. The network takes the RGB image of the object as the input. The RGB image of the object is fed into a feature extraction layers based on ResNet-50 [16]. We use the first 4 stage of the ResNet-50 as the feature extraction layers. The input image is resized to be a fixed size of 448×448 as the network's input. So, the feature extraction layers will output the feature map with a size of 28 × 28.

With the coordinates of semantic regions from the detector's result (in testing stage) or ground truth (in training stage), we slice the corresponding regions from the feature map outputted by the feature extraction layers and normalize their size to 7X7 by a RoIAlign operation. For each kind of semantic regions, different sub-classifiers is trained. Those sub-classifiers have the same architecture. We design the architecture of sub-classifier follow ResNet [16], which contains residual block, pooling layer, and fully connected layers. Each semantic region is fed to the corresponding sub-classifier and the output of a sub-classifier is a $c$-dimensional vector generated by a $c$-way fully connected layer with softmax nonlinear activation function. $T$ sub-classifiers will output $T$ vectors. The $i$-th sub-classifier's output vector can be written as $h^i, h^i, , h^i, h^i$, where $h^i$ represent the confidence of $i$-th segamtic region belongs to the $j$-th categories. In order to comprehensively consider every semantic region, all the sub-classifiers are combined to generate a stronger classifier. We fuse the result of each sub-classifier to get the stronger classifier's result by a plurality voting strategy. The final fine-grained classifier result of the object is got by the equation

$$J = \mathop{\mathrm{argmax}}_{j} \sum_{i=1}^{T} h_j^i \qquad (1)$$

$J$ represents the classification result that the object belongs to the $J$-th category.

In the training stage, we train the whole classifiers' network by optimizing the sum of all the cross-entropy cost of sub-classifier's prediction and groundtruth lables. The whole classifier network's loss function can be written as

$$Loss = \sum_{i=1}^{T} l_i, \quad (2)$$

where

$$l_i = -\sum_{j=1}^{C} t_j \ln h_j^i. \quad (3)$$

$l_i$ is the cross-entropy loss for the *i*-th sub-classifier. $t_j$ is the one-hot label which represents the object's category. Though the network contains different sub-classifiers, they are trained simultaneously by minimum a single objective function, making the classifier's training stage to be an end- to-end process. The classification module's whole architecture is designed by a share-and-divide strategy. The whole network shares the same feature extraction layers. Shared feature extraction layers can improve computational efficiency and reduce the number of parameters. Sub-classifiers have the same architecture but do not share parameters. Special sub-classifier for special semantic region makes the sub-classifier more professional. More important, ensemble learning aims to combine multi weaker classifiers and to get a stronger classifier. The diversity between multi-classifier is the necessary condition to ensure the total classifier be stronger and more accurate. In our frame- work, different sub-classifiers have different semantic regions as input and are trained professional for the special kind of semantic region. It makes those sub-classifiers naturally contain diversity. Combining them will get a more accurate classifier and relieve overfitting which makes the classifier stronger.

### C. Data Augmentation

When training a deep neural network, data augmentation is an important technology that can artificially enlarge the dataset using transformations. When training the faster R-CNN based detector, we use data augmentation method following [10]. When training the classification network, in addition to using random flip, crop and rotation data augmentation methods, we also propose a data augmentation method which not only can enlarge dataset but also can improve the classifier's robustness about the detection result. In the classifier's training stage, groundtruth coordinate of the semantic regions is used for RoIAlign operation. To implement data augmentation, we adjust the coordinate of semantic regions to increase the dataset. Center shifting and size scaling is the two method to adjust the coordinate. Hypothesizing the coordinates of a semantic part bounding box is *x, y, w, h*, where *x, y* represent the center of the bounding box and *w, h* denote the width and height of bounding box. The coordinate of the whole object bounding box is $x_o, y_o, w_o, h_o$. Center shifting changes the center coordinate *x, y* by the equation

$$x' = x + \alpha x_o, \quad (4)$$
$$y' = y + \beta y_o, \quad (5)$$

where α and β are random variables following a Gaussian distribution whose expectation is 1 and standard deviation is 0.1. If the absolute value of α or β is larger than the distribution's standard deviation, it will be discard and repicked. Size scaling adjust the size of the semantic region's bounding box by the equation

$$w' = \gamma w, \quad (6)$$
$$h' = \delta h, \quad (7)$$

where, γ and δ follow a Gaussian distribution whose expectation is 1.1 and the standard deviation is 0.2. γ or δ will also be dropped and repicked if its magnitude is more than one standard deviations from the expectation. After the implementation of center shifting and size scaling, the new semantic part bounding box with coordinate *x', y', w', h'* is got. Those adjustments make the regions for training have some offset from the origin oracle region annotations. This not only enlarges the dataset but also improves the classifier's robustness about the detection result. Because in the testing stage, the detection result often has offset from the origin oracle region annotations. If the offset is considered in the training stage, the classifier will be more robust in the testing stage.

### III. EXPERIMENT

#### A. CUB-2011 Dataset

We test our approach on a widely used dataset for fine- grained classification, Caltech-UCSD birds [4] (CUB-2011). CUB-2011 is a bird dataset that contains 200 bird species. Each bird species has about 60 images with labels. The label of each image contains the bird species, the bounding box of the whole bird object and the coordinates of 15 kinds semantic part center such as bird's back, belly, tail, right and left legs and so on. The width and height of the semantic region bounding box are set as $\frac{W}{2}$ and $\frac{H}{2}$, where *W* and *H* represent the detector,

#### B. Implementation Details

For the detection module, we set the hyper-parameters following [10]. The ResNet-50 is used as the detector's backbone. The detector's network is pre-trained on the PASCAL VOC 2012 dataset [17] with 20 kinds objects and we implement fine-tune for the CUB-2011 dataset. In the classifier's training stage, we initialize the feature extraction layers using weights from ResNet-50 pre-trained on ImageNet [18] for 1000-class recognition and initialize the sub-classifiers' weights using random normal initializer. We use Adam optimizer. $L^2$ regularization is used to relieve overfitting. In our experiments, a GTX 1070 Ti GPU machine and publicly available Keras framework is used for training and testing.

TABLE I
THE AP AND MAP OF THE DETECTION RESULT.

| Semantic Region | Average Precision |
|---|---|
| head | 93.8% |
| back | 92.6% |
| belly | 91.5% |
| wing | 90.1% |
| leg | 89.3% |
| tail | 93.1% |
| breast | 90.4% |
| mAP | 91.5% |

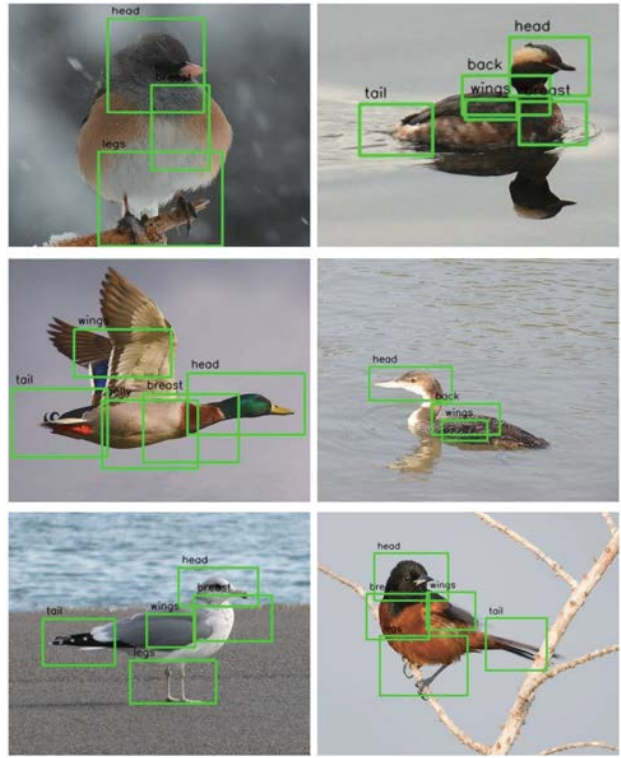

Fig. 2. Some detection result.

TABLE II
THE ACCURACY OF EVERY SUB-CLASSIFIER.

| Sub-classifier | Accuracy |
|---|---|
| head | 72.3% |
| back | 56.8% |
| belly | 53.4% |
| wing | 69.4% |
| leg | 45.6% |
| tail | 44.7% |
| breast | 60.4% |
| Final Accuracy | 80.3% |

*C. Detection Result on CUB-2011 Dataset*

Because the detector's result is the input of the classifier, the high quality of detection result is the necessary condition for accurate recognition. We evaluate our detector by calculating the Average Precision (AP) for every kind of semantic regions and the mean Average Precision (mAP) on the test dataset. The AP and mAP are shown in Tab. I, where the 2-nd to 8-th lines represent the AP for each semantic region and the last line of the table represents the mAP. This result shows our faster R- CNN based detector have a good result. Some detection result of semantic regions is visualized in Fig. 2.

*D. Classification Accuracy on the CUB-2011 Dataset*

We evaluate the classifier's quality on the test dataset of CUB-2011. Both the sub-classifiers' accuracy and the accuracy of the framework's final result are calculated and are shown in Tab. II. As shown in Tab. II, the accuracy of the sub- classifier is ranged from $44.7\%$ to $72.3\%$. But the accuracy of the framework's final result is $80.3\%$, which is higher than any sub-classifier's accuracy. The reason is that every sub-classifier considers only one semantic region, combining all the sub- classifiers' result will get a stronger classifier. This proves the effectiveness of our ensemble learning method. Tab. III shows the comparison of accuracy between our approach and other approaches. As shown in Tab. III, our approach achieves a good result.

*E. Remote Scene Recognition in NWPU-RESISC45*

Our approach is designed for fine-grained classification, but it can also be used for remote scene recognition. We test it on the NWPU-RESISC45 dataset [14]. NWPU-RESISC45 is a publicly available dataset for REmote Sensing Image Scene Classification(RESISC), which contains 45 different classes of remote scene and 700 images in each class. In our experiment, we select four classes of them, which are harbor, parking lot, church and thermal power station. Some example images are shown in Fig. 3. Just like fine-grained classification, the discriminative features of the remote scene often appear in semantic regions, such as vehicles in the parking lot, ships in the harbor, hyperbolic cooling towers of the thermal power station or domes of the church. We use those object as the semantic regions and evaluate our model on the four classes of the scene. We get a good accuracy of $86.3\%$. Our approach is proved useful for remote scene recognition.

TABLE III
COMPARISON WITH DIFFERENT METHODS ON CUB-2011.

| Method | Accuracy |
|---|---|
| Lin *et al.* [19] | 80.3% |
| Zhang *et al.* [8] | 76.4% |
| Goring *et al.* [20] | 57.8% |
| Huang *et al.* [9] | 76.6% |
| Lin *et al.* [21] | 80.4% |
| Ours | 80.3% |

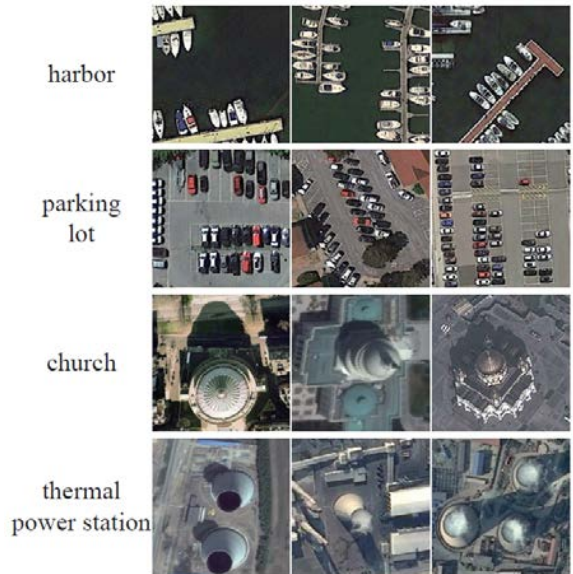

Fig. 3. NWPU-RESISC45 dataset example.

CONCLUSION

In this paper, we introduce a novel region-based ensemble learning method for fine-grained classification. Coordinates of semantic regions of the objects are first located by a faster R-CNN based detector. The classification module uses an ensemble learning method. With different semantic regions, different sub-classifiers are trained. Finally, those sub-classifiers are combined by a plurality voting strategy and a stronger classifier is got. The classification module is designed by a share-and-divide strategy, each sub-classifier share the same feature extraction layers and has owns parameters. This design is not only computational efficient but also can guarantee diversity and professionalism of sub-classifiers. We test our approach on the CUB-2011 dataset and our model is proved to be accurate. We also extend our approach for remote scene classification and get a good result.

## ACKNOWLEDGMENT

This work is partially supported by the National Key R&D Program of China (2016YFE0204200), the National Natural Science Foundation of China (61503017, U1435220, 61703287), the Aeronautical Science Foundation of China (2016ZC51022), Natural Science Foundation of the Higher Education Institutions of Jiangsu Province of China (18KJB520021), the Open Research Fund of Fujian Engineering Research Center of Public Service Big Data Mining and Application, Fuzhou, China.